\documentclass{bmvc2k}
\usepackage{multirow}
\usepackage{color}
\usepackage{booktabs}
\usepackage{colortbl}

\title{Hierarchical Graph Networks for\\ 3D Human Pose Estimation}

\addauthor{Han Li}{qingshi9974@sjtu.edu.cn}{1}
\addauthor{Bowen Shi}{sjtu_shibowen@sjtu.edu.cn}{1}
\addauthor{Wenrui Dai$^\ast$}{daiwenrui@sjtu.edu.cn}{1}
\addauthor{Yabo Chen}{chenyabo@sjtu.edu.cn}{1}
\addauthor{Botao Wang}{botaow@qti.qualcomm.com}{2}
\addauthor{Yu Sun}{sunyu@qti.qualcomm.com}{2}
\addauthor{Min Guo}{mguo@qti.qualcomm.com}{2}
\addauthor{Chenlin Li}{lcl1985@sjtu.edu.cn}{1}
\addauthor{Junni Zou}{zoujunni@sjtu.edu.cn}{1}
\addauthor{Hongkai Xiong}{xionghongkai@sjtu.edu.cn}{1}

\addinstitution{
 Shanghai Jiao Tong University\\
 Shanghai, China
}
\addinstitution{
 Qualcomm AI Research\\
 Shanghai, China
}

\runninghead{Han Li \etal}{HGN for 3D Human Pose Estimation}

\def\eg{\emph{e.g}\bmvaOneDot}

\def\etal{\emph{et al}\bmvaOneDot}
\begin{document}
\abovedisplayshortskip=0pt
\belowdisplayshortskip=0pt
\abovedisplayskip=0pt
\belowdisplayskip=0pt
\maketitle

\begin{abstract}
Recent 2D-to-3D human pose estimation works tend to utilize the graph structure formed by the topology of the human skeleton. However, we argue that this skeletal topology is too sparse to reflect the body structure and suffer from serious 2D-to-3D ambiguity problem. To overcome these weaknesses, we propose a novel graph convolution network architecture, Hierarchical Graph Networks (HGN). It is based on denser graph topology generated by our multi-scale graph structure building strategy, thus providing more delicate geometric information. The proposed architecture contains three sparse-to-fine representation subnetworks organized in parallel, in which multi-scale graph-structured features are processed and exchange information through a novel feature fusion strategy, leading to rich hierarchical representations. We also introduce a 3D coarse mesh constraint to further boost detail-related feature learning. Extensive experiments demonstrate that our HGN achieves the state-of-the-art performance with reduced network parameters. Code is released at
 \href{https://github.com/qingshi9974/BMVC2021-Hierarchical-Graph-Networks-for-3D-Human-Pose-Estimation}{https://github.com/qingshi9974/BMVC2021-Hierarchical-Graph-Networks-for-3D-Human-Pose-Estimation.}


\end{abstract}

\section{Introduction}
\label{sec:in}
3D human pose estimation aims to predict the 3D spatial coordinates of body joints from a monocular image and has been widely exploited in various applications such as abnormal behavior detection, sports analysis and automated driving. In recent years, 2D human pose estimation performance has been greatly improved owing to more refined network structure design and richer 2D human pose datasets. Recent works show that using such detected 2D joints positions, the 3D human pose can also be efficiently and accurately regressed~\cite{pavllo20193d,2017simple,2020Deep,fang2018learning,cheng2019occlusion}. To further boost performance, many attempts have been made to explicitly utilize the  human skeletal topology and use graph convolutional networks (GCNs) to exploit the spatial configurations for 3D human pose estimation~\cite{2019Exploiting,zhao2019semantic,2021Graph,liu2020comprehensive}. However, the graph topology of human skeleton is usually formed by few number of joints (\eg 17 joints in Human3.6M~\cite{ionescu2013human3} dataset) and is sparse. In this paper, we raise a critical issue: are such few number of joints enough for reflecting the body structure and estimating 3D human pose?

We analyze the drawbacks of sparse graph representation from two perspectives. First, regressing the 3D human pose based on 2D joints positions is an ill-posed problem since multiple valid 3D poses can be projected to the same 2D pose. This inherent ambiguity problem will be more serious when the defined human skeleton structure is oversimple, thus hindering performance improvement. Second, each part of the human body is sparsely represented by only one joint, which will impede the expression of local information and lead to positioning failure when facing complex motions and occluded scenes.  

\begin{figure}[!t]
\vspace{-3pt}
\renewcommand{\baselinestretch}{1.0}
\setlength{\abovecaptionskip}{-10pt}
\setlength{\belowcaptionskip}{-10pt}
\centering
\includegraphics[width=11.5cm]{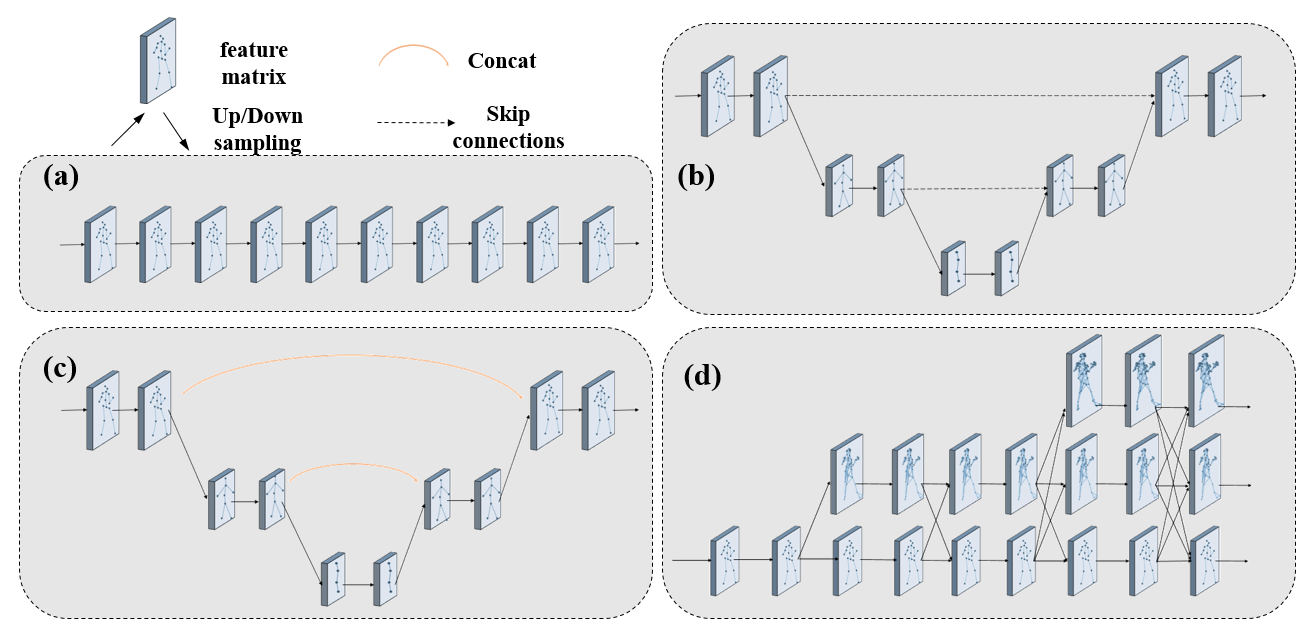}
\caption{Several typical GCN architectures for 2D-to-3D human pose estimation: (a) Straight-forward architecture~\cite{zhao2019semantic}. (b) Graph Stacked Hourglass~\cite{2021Graph}  (c) Graph U-Nets~\cite{gao2019graph}. (d) Ours Hierarchical Graph Networks. (b) and (c) also leverage multi-scale features but their largest scale graph is formed by human skeleton topology, while our architecture introduces denser graph topology and organizes the network in a sparse-to-fine way. Note that our architecture may look like HR-Net~\cite{sun2019hr} but the network organization mode and feature fusion strategy are totally different.   
} \label{fig:various}
\end{figure}

In this paper, we address these issues by exploiting denser graph topology,
proposing a novel architecture named \textit{Hierarchical Graph Networks (HGN)}. Specifically, we propose a novel graph structure building strategy that utilizes the human shape information to obtain finer human structure representation. Then, starting from a sparse representation subnetwork, we gradually add sparse-to-fine representation subnetworks, and connect the multi-scale subnetworks in parallel. Since the 
mapping relation between sparse and fine representations is hard to model by manually designed  graph pooling and unpooling, 
we also propose a multi-scale feature fusion strategy to learn the suitable mapping and exchange information across the parallel subnetworks . In Figure~\ref{fig:various}, we illustrate several typical GCN architectures for 2D-to-3D human pose estimation. The advantage of our HGN is that more delicate features are able to be extracted from the sparse-to-fine human structures. 

Though promising it is, merely increasing the complexity of the graph structure without giving meaning to its nodes can only bring a limited performance improvement. Inspired by recent dense mesh vertices estimation methods~\cite{choi2020pose2mesh}, we conduct dense vertices coarsening to obtain the pseudo-groundtruth of the coarse mesh vertices and utilize it as an additional constraint. Since sparse mesh can represent the shape information of human body, it contains more abundant and detailed information. Adding additional mesh constraints makes our model extract detail-related features, which is of great help to the evaluation of some joints with high degrees of freedom. 


In a nutshell, this paper makes the following contributions:
\begin{itemize}
\item We build a novel hierarchical graph network with multi-scale feature fusion. It is based on denser graph topology generated by our multi-scale graph structure building strategy and contains more delicate geometric information.  
\item We generate coarse mesh vertices pseudo-groundtruth by dense vertices coarsening and utilize it as an additional constraint to make the model pay more attention to detailed information.
\item Experiment results demonstrate the superior performance of our HGN compared to other state-of-art GCN-based methods. The key idea of designing more delicate human structure representation may shed light on future research direction.
\end{itemize}

\section{Related Work}
\textbf{3D Human pose estimation.} Current 3D human pose estimation can be categorized into two types: one-stage approach and two-stage approach. The one-stage approach takes  RGB images as input for 3D pose estimation.  
With the rapid development of deep learning, 
recent works~\cite{zhou2017towards,shi2020tiny,2018Ordinal,moon2019camera,lin2020hdnet,sun2017compositional} take advantages of Convolutional Neural Networks (CNNs)  for the image-to-3D human pose estiamtion.
Zhou \etal~\cite{zhou2017towards} propose  a weakly-supervised transfer learning method to make full use of mixed 2D and 3D labels, which augments the 2D pose estimation sub-network with a 3D depth regression sub-network to regress the depth. Sun \etal~\cite{sun2018integral} employ soft-argmax operation to regress the 3D coordinates of body joints in a differentiable way. Pavlakos \etal~\cite{2018Ordinal} exploit voxel to discretize representations of the space around the human body and use 3D heatmaps to estimate 3D human pose.

The  second category of approaches breaks the problem down into two steps: first predicting 2D human joints from the input image, and then lifting 2D joints  to predict 3D pose. Our approach falls into this category. Martinez \etal~\cite{2017simple} propose a simple yet effective baseline for 3D human pose estimation, it uses only 2D joints as input but gets highly accurate results, showing the importance of 2D joints information for 3D human pose estimation. Since the  skeleton’s topology can be viewed as a graph structure, there has been increasing use of Graph Convolutional Networks (GCNs) for 2D-to-3D  pose estimation tasks~\cite{zhao2019semantic,liu2020comprehensive,2019Exploiting,2021Graph,kong2020sia,zeng2021learning}.

\textbf{Graph Convolutional Networks.} In recent years,
Graph Convolutional Networks (GCNs) have been widely used to process graph-structured data, it can be regarded as a generalization of traditional CNNs. In general, GCNs can be divided into two categories: spectral GCN~\cite{defferrard2016convolutional,bruna2013spectral} and spatial GCN~\cite{niepert2016learning,kipf2016semi,ve2018graph,graphsage}, in which our approach falls into the second category. In the early day, Kipf and Welling~\cite{kipf2016semi} introduce the "vanilla" GCN,  which performs the transformation and aggregation of graph-structured data via a simple graph convolution. 
Based on "vanilla" GConv, 
Zhao \etal~\cite{zhao2019semantic}  propose Semantic Graph Convolution (SemGConv), which can learn local and global semantic relations among nodes
in the graph by adding a parameter matrix. 
Zou \etal~\cite{zou2020high} exploit a high-order GCN to learn long-range dependencies among body joints. 
However, they all adopt a straight-forward network architecture which simply stacks residual graph convolution blocks with same graph topology. 
To extract multi-scale features, Cai \etal~\cite{2019Exploiting} designed a U-nets like graph networks architecture and Xu \etal~\cite{2021Graph} proposed Graph Stacked Hourglass Networks. Unlike the above methods, we argue that the human skeletal graph used in these works is too sparse and propose hierarchical graph networks by exploiting denser graph topology.


\section{Method}
\subsection{Preliminaries}
\label{sec:gcn}
Let $\mathcal{G}=\left\{\mathcal{V}, \mathcal{E}\right\}$ denote a graph where $\mathcal{V}$ is a set of N  nodes and  $\mathcal{E}$  is the collection of edges. $\mathbf{A}\in\{0,1\}^{N\times N}$ is the adjacency matrix of $\mathcal{G}$, and we have $a_{ii}=1$ and $a_{ij}=1$ if $j$ is the neighboring node of $i$. Each node $i$ is associated with a $D$-dimensional feature vector $x_i\in \mathcal{R}^D$, and the node representations are collected into a matrix $\mathbf{X}\in \mathcal{R}^{D\times N}$. The vanilla GCN~\cite{kipf2016semi} is\begin{equation}\label{eq:vanilla-gcn}
\mathbf{X}^{\prime}=\sigma\left(\mathbf{W}\mathbf{X}\mathbf{\Tilde{A}}\right)\text{,}
\end{equation}
where $\mathbf{X}^{\prime}\in \mathcal{R}^{D^{\prime}\times N}$ is the updated feature matrix and $\sigma(\cdot)$ is a non-linear function. $\mathbf{\Tilde{A}}$ denotes a symmetrically normalized version of $\mathbf{A}$ and $\mathbf{W}\in\mathcal{R}^{D^{\prime}\times D}$ is a learnable matrix that transforms node representations.  SemGConv~\cite{zhao2019semantic} further learns the semantic relationships of neighboring nodes by adding another learnable weighting matrix $\mathbf{T}\in\mathcal{R}^{N\times N}$.
\begin{equation}\label{eq:Semgconv}
\mathbf{X}^{\prime}=\sigma\left(\mathbf{W}\mathbf{X}\rho_i\left(\mathbf{T}\odot\mathbf{\Tilde{A}}\right)\right)\text{,}
\end{equation}
where $\odot$ is an element-wise product operation and $\rho_i(\cdot)$ is softmax nonlinearity which normalizes the input matrix across all choices of $i$. Following previous works~\cite{liu2020comprehensive,zhao2019semantic,yan2018spatial}, we use two different transformation matrices for the representation of each node $i$ and its neighbors respectively in actual implementation.
\begin{figure}[!t]
\renewcommand{\baselinestretch}{1.0}
\setlength{\abovecaptionskip}{-15pt}
\setlength{\belowcaptionskip}{-10pt}
\begin{center}
\includegraphics[width=11.5cm]{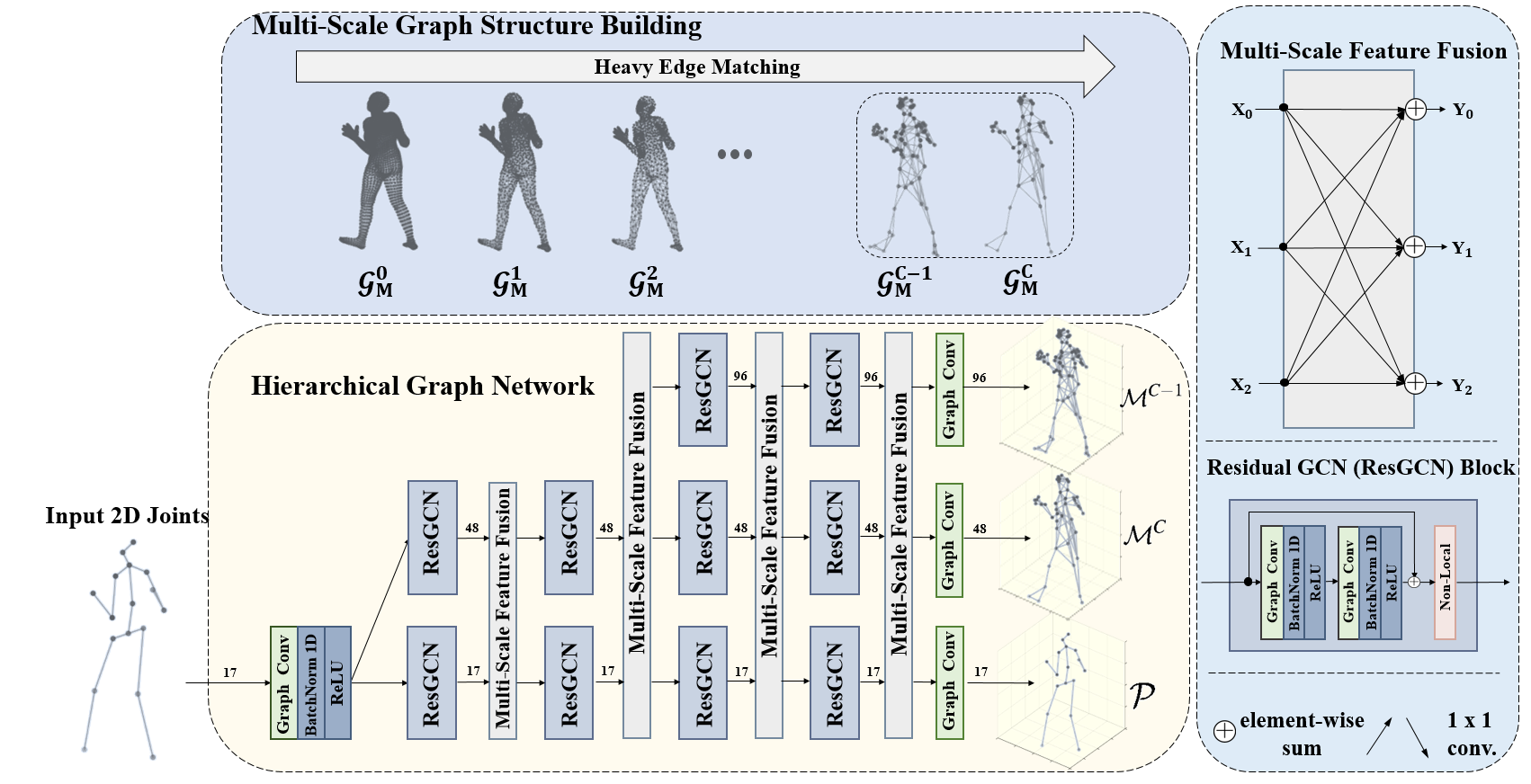}
\end{center}
\caption{The overall pipeline of HGN. The input 2D joints $\mathbf{J}$ passes three subnetworks of HGN to obtain multi-scale features and output the predicted 3D Pose $\mathcal{P}$ and coarse mesh $\mathcal{M}^{C}$ and $\mathcal{M}^{C-1}$ in parallel. The nodes of graph built for three subnetworks from bottom to up is 17, 48 and 96, respectively, using the multi-scale graph structure building processing shown in the upper part. The down-right part illustrates the Residual GCN (ResGCN) block and the up-right part shows the proposed multi-scale feature fusion strategy used in HGN. }\label{fig:pipeline}
\end{figure}

\subsection{Hierarchical Graph Networks}

As mentioned in Section~\ref{sec:in}, the 2D-to-3D human pose estimation is an ill-posed problem, and it will become more serious when occlusion occurs. We believe that the above problem can be alleviated if we leverage denser graph structures to depict the human skeleton, thus increasing the complexity of the constraint relationship between human joints. To achieve this goal, as shown in Figure~\ref{fig:pipeline}, we propose Hierarchical Graph Networks (HGN) consisting of three subnetworks with different scales of graph structure. Novel graph structure building strategy and multi-scale feature fusion strategy are also designed for HGN. 

{\bf Multi-scale graph structure building.} We build three different human structure graphs and assign them to the three subnetworks of HGN in a sparse-to-fine way. The sparsest graph for the bottom subnetwork in Figure~\ref{fig:pipeline} is defined on the standard human skeletal graph $\mathcal{G}_P=\left\{\mathcal{V}_P, \mathcal{E}_P\right\}$. Similar to existing methods~\cite{2017simple,zou2020high,liu2020comprehensive}, the number of nodes $J$ is set to 17 in this paper. For the middle and top subnetworks, we coarsen the human mesh graph $\mathcal{G}_M=\left\{\mathcal{V}_M, \mathcal{E}_M\right\}$, which represents human shape information and contains 6890 vertices, multiple times using Heavy Edge Matching (HEM)~\cite{defferrard2016convolutional} and obtain a set of various scales of graphs $\{\mathcal{G}_M^{c}\}_{c=0}^C$, where $c$ indicates the coarsening level. Although more complex graphs can be allocated, we choose the two coarsest graphs $\mathcal{G}_M^{C-1}$ and $\mathcal{G}_M^{C}$ whose the number of nodes $V^c$ equal to 96 and 48, respectively, to maintain a suitable model size.


{\bf Multi-scale feature fusion.} 
Since the human skeletal graph $\mathcal{G}_P$ and the coarse human mesh graphs $\mathcal{G}_M^{C-1}$ and $\mathcal{G}_M^{C}$ are constructed in different ways, it is difficult to describe the specific corresponding relationship between them in a manual way or simply graph downsampling and upsampling operations \cite{2019Exploiting,2021Graph}. As an alternative, we design a multi-scale feature fusion strategy to learn their mapping relations.

Inspired by the exchange units proposed by Sun \etal~\cite{sun2019hr}, we conduct multi-scale fusions  such that the  parallel subnetworks can exchange different scales information from each other. Given the input feature  with different number of nodes $\{\mathbf{X_0},\mathbf{X_1},\ldots,\mathbf{X_s}\}$, where the subscript denotes the graph scale, we can obtain the fused feature $\{\mathbf{Y_0},\mathbf{Y_1},\ldots,\mathbf{Y_s}\}$ whose scale
and widths are the same to the input by
\begin{equation}
    \label{eq:fusion}
    \mathbf{Y}_k = \sum_{i=1}^s\operatorname{a}(\mathbf{X}_i,k)\text{,}~~ k=0,1,\ldots,s,
\end{equation}
where function $\operatorname{a}(\mathbf{X}_i,k)$ 
consists of upsampling or downsampling $\mathbf{X}_i$ from graph scale  $i$ to scale $k$.  We use $1\times 1$ convolution for both upsampling and downsampling since the mapping relations are proved to be learned well in such a simple way.


{\bf Network architecture.}
As shown in Figure~\ref{fig:pipeline}, the input 2D pose is mapped to the latent space by a pre-processing graph convolution layer. Then, the multi-scale features extracted in the three subnetworks 
are fused repeatedly. The subnetworks have 4, 4, 2 residual blocks~\cite{he2016deep}, respectively from bottom to up. Each residual block consists of two graph convolution layers and is followed by a non-local layer~\cite{wang2018non} to capture both local and global information, and all the graph convolution layers are followed by batch normalization~\cite{ioffe2015batch} and a ReLU non-linear layer~\cite{nair2010rectified}.  Finally, the features are fed into the output convolution layer and mapped to the output space. It is noted that our HGN does not rely on a specific graph convolution method, so both SemGConv and Vanilla GConv introduced in Section~\ref{sec:gcn} can be implemented in our architecture. 


\subsection{Mesh Constraint}
Since graphs $\mathcal{G}_M^{C-1}$ and $\mathcal{G}_M^{C}$ are derived from mesh graph $\mathcal{G}_M$, we can also generate coarse mesh vertices pseudo-groundtruth by dense vertices coarsening and leverage it as another constraint to further refine the local feature representation. Specifically, we generate 3D human mesh $\mathbf{M}$ with groundtruth vertices location by fitting SMPL parameters to the 3D groundtruth poses using SMPLify-X~\cite{bogo2016keep}, and then obtain the pseudo-groundtruth of two coarsest meshes $\mathbf{M}^{C-1}\in\mathcal{R}^{V_{C-1}\times 3}$  and  $\mathbf{M}^{C}\in\mathcal{R}^{V_C\times 3}$ by a pre-defined indices mapping operation. The final loss function is a combination of 3D pose and 3D coarse mesh constraint:\begin{equation}
\mathcal{L}(\mathcal{P},\mathcal{M}^{C-1},\mathcal{M}^{C})=\lambda_{P}\underbrace{\sum_{i=1}^J\|\hat{\mathbf{p}}_i-\mathbf{p}_i\|^2_2}_{\text{3D pose loss}}+\lambda_{M}\underbrace{\left(\sum_{i=1}^{V_{C-1}}\|\hat{\mathbf{m}}^{C-1}_{i}-\mathbf{m}^{C-1}_{i}\|^2_2+\sum_{i=1}^{V_{C}}\|\hat{\mathbf{m}}^{C}_{i}-\mathbf{m}^{C}_{i}\|^2_2\right)}_\text{3D coarse mesh loss}\text{,}
    \label{eq:loss}
\end{equation}
where $\lambda_{P}=1$, $\lambda_{M}=0.01$. $\mathcal{P}=\left\{\mathbf{\hat{p}}_i|i=1,\ldots,J\right\}$ are the predicted 3D pose and  $\mathcal{M}^c=\left\{\mathbf{\hat{m}^c}_i|i=1,\ldots,V_c,~~c=C-1,C\right\}$ are the predicted 3D coarse meshes. $\mathbf{p}_i$  and $\mathbf{m}^c_i$ are the groundtruth /pesudo-groudtruth corresponding to $\mathbf{\hat{p}}_i$ and $\mathbf{\hat{m}^c}$.
\section{Experiments}
\label{sec:exp}
\subsection{Experimental Setups}
{\bf Dataset.}  We mainly evaluate our proposed method on the Human3.6M dataset~\cite{ionescu2013human3}, which is widely used in the 3D human pose estimation task. It provides 3.6 million color images taken from four synchronous cameras in different positions and perspectives, by
recording 11 subjects actors performing 15 different actions, such as eating and walking. 
There are 7 subjects annotated with 3D joints. For fair comparison, we follow previous works~\cite{2017simple,zhao2019semantic,2021Graph} and choose 5 subjects (S1, S5, S6, S7, S8) for training and the other 2 subjects (S9, S11) for test. Besides, the MPI-INF-3DHP~\cite{mehta2017monocular} test set  provides images in three different scenarios: studio with
a green screen (GS), studio without green screen (noGS) and outdoor scene (Outdoor). 
We apply our model to this dataset to test the generalization capabilities of our proposed method.


{\bf Evaluation.} For the Human3.6M benchmark, there are two evaluation protocols used in previous works~\cite{2017simple,zhao2019semantic,2021Graph}. Protocol \#1 uses the mean per-joint position error (MPJPE) as evaluation metric, which
computes the mean Euclidean distance error per-joints between the predicted 3D joints and the ground truth after the origin (pelvis) alignment. Protocol \#2 aligns the predicted 3D joints with the ground truth by rigid transformation and then computes the error. This metric is abbreviated as PA-MPJPE. Both of these two metrics are measured in millimeter (mm).

{\bf Implement details.}
We implement our method within the PyTorch framework. During the training stage, we choose the Adam optimizer~\cite{kingma2014adam} with the learning rate initialized to 0.001 and decayed by 0.9 per 20 epochs. We train each model for 100 epochs using a mini-batch size of 64. We initialize weights of the GCNs using the initialization method described in~\cite{glorot2010understanding}. To avoid overfitting, we also adopt Max-norm regularization. In the following experiment, unless specified, the SemGConv is used as the graph convolution layer.

\subsection{Ablation Study}
\label{sec:ablation}
We conduct a series of ablation studies to better understand how each component affects the performance. The 2D ground truth is taken as input. 

{\bf Effects of mesh constraint.} We first diagnose how the mesh constraint affects the performance. As shown in Table~\ref{table:abla-on-loss}, our method achieves the best performance when setting the weight of coarse mesh constraint $\lambda_{M}$ to 0.01. Compared with not adding mesh constraint ($\lambda_{M}=0$), we achieve 0.99mm (38.31 to 37.32) and 0.53mm (29.04 to 28.51) gain under two protocols respectively. The performance gain is mainly due to that the mesh constraint enriches the fine-grained representation. However, if we increase $\lambda_{M}$, the performance dramatically drops. We believe that this is because the pose-related information will be covered by shape-related information using higher mesh constraint weight. Therefore, it makes sense to set a small weight for mesh constraint for better pose estimation.

\begin{table}[!t]
\renewcommand{\baselinestretch}{1.0}
\renewcommand{\arraystretch}{1.0}
\centering
\small
\caption{Ablation study on effects of mesh constraint. Different weights of mesh constraint $\lambda_{M}$ in Equation~\eqref{eq:loss} are set and  $\lambda_{M}= 0$  means removing the mesh constraint.}\label{table:abla-on-loss}
\begin{tabular}{l|l|c|c}
\hline
$\lambda_{P}$&$\lambda_{M}$&MPJPE (mm)&PA-MPJPE (mm)\\ \hline\hline
1 & 0 &38.31 &29.04 \\ \hline
1 & 0.001 &37.68 &28.79\\ \hline
1 & 0.01 &{\bf37.32} &{\bf 28.51} \\ \hline
1 & 0.1 &38.69 &29.09 \\ \hline
1 & 1 & 39.64 &29.81 \\ \hline
\end{tabular}
\end{table}
\begin{table}[!t]
\renewcommand{\baselinestretch}{1.0}
\renewcommand{\arraystretch}{1.0}
\centering
\small
\caption{Ablation study on effects of denser graph topology with $\lambda_M$ set to 0.01.}  
\begin{tabular}{l|cc|cc|c}
\hline

\multirow{2}{*}{Method} & MPJPE& PA-MPJPE& \multicolumn{2}{c|}{MPVPE}& \multirow{2}{*}{\# Params}\\
\cline{4-5}
&(mm)&(mm)&$\mathcal{M}^{C-1}$&$\mathcal{M}^{C}$ \\
\hline
\hline
SemGCN (Baseline)~\cite{zhao2019semantic} &40.78&31.01 &-&-&0.43M  \\ \hline
HGN w/o $\mathcal{G}_M^{C}$ & 38.69&29.07 &66.28&-&0.82M  \\ \hline
HGN w/o $\mathcal{G}_M^{C-1}$ &37.83 &28.71&-&65.92&0.81M \\ \hline
HGN  &{\bf37.32} &{\bf 28.51}    &62.74&61.76&1.04M \\ \hline
\end{tabular}
\label{table:abla-on-network-structure}
\vspace{-15pt}
\end{table}

{\bf Effects of denser graph topology.} We then inspect how denser graph topology benefits the representation.We first set $\lambda_M$ to 0.01 and carry out experiments with three variants of our HGN. 1) {\bf SemGCN (Baseline)}: only the bottom subnetwork in Figure~\ref{fig:pipeline} defined on  $\mathcal{G}_P$ is reserved. This straight-forward architecture is equivalent to Semantic Graph Convolutional Networks (SemGCN)~\cite{zhao2019semantic}, and we treat it as our baseline. 2) {\bf HGN w/o $\mathcal{G}_M^{C}$}: we remove the top subnetwork and change the graph structure in the middle subnetwork from $\mathcal{G}_M^{C}$ to $\mathcal{G}_M^{C-1}$, so that the network can output 3D pose and coarse mesh $\mathcal{M}^{C-1} $ containing 96 vertices in parallel. 3) {\bf HGN w/o $\mathcal{G}_M^{C-1}$}: we remove the top subnetwork and do not modify the other subnetworks to output 3D pose and coarse mesh $\mathcal{M}^{C} $ containing 48 vertices. The results are shown in Table~\ref{table:abla-on-network-structure}, in which the mean per vertex position error (MPVPE) for mesh prediction measurement is also listed as a reference. We find that all networks with delicate graph structures outperform the baseline for a large margin, which proves the benefits of introducing denser graph topology. Our HGN achieves the best results in MPJPE (37.32mm), indicating that our proposed model has a strong ability to leverage sparse-to-fine graph structures.

Furthermore, we set $\lambda_M$ to 0 to remove the influence of mesh constraint and model parameters. Experiments are made by fixing the number of channels and model parameters, respectively. Table~\ref{table:compare with GSH_control_channel} shows that our HGN still achieves the overall best performance.

{\bf Effects of hierarchical network structures.}
To verify that our architecture has better performance than some typical structures, we keep the GCN type same and the model size comparable for fair comparison. Table~\ref{table:compare with GSH} shows that the model using our architecture performs better than Sequential Residual blocks (denoted as SeqRes)~\cite{zhao2019semantic} and Graph Stacked Hourglass (denoted as GraphSH)~\cite{2021Graph} architecture with fewer parameters. It is noted that our method boosts the performance for a large margin when using traditional Vanilla GConv, which demonstrates the great advantage of our architecture itself. 




\begin{table}[!t]
\renewcommand{\baselinestretch}{1.0}
\renewcommand{\arraystretch}{1.0}
\setlength{\abovecaptionskip}{-2pt}
\small
\centering
\caption{Ablation study on the effect of denser graph topology with $\lambda_M$ set to 0. The number of channels and model parameters are fixed (to 128 and 0.43M) for evaluations, respectively. }
\begin{tabular}{l|c|c|c|c}
\hline
Method &\# Channels&MPJPE (mm)&PA-MPJPE (mm)&\# Params \\ \hline\hline
SemGCN (Baseline)~\cite{zhao2019semantic}&128& 40.78 &31.01&0.43M \\\hline
HGN w/o $\mathcal{G}_M^{C}$ &128& 38.94& 29.47& 0.82M \\  \hline
HGN w/o $\mathcal{G}_M^{C-1}$ & 128&38.59&29.18&0.81M\\ \hline
HGN & 128&{\bf38.31}& {\bf29.04}& 1.04M \\ \hline
\hline
HGN w/o $\mathcal{G}_M^{C}$ &91&39.87&30.31&0.43M \\  \hline
HGN w/o $\mathcal{G}_M^{C-1}$&92&39.71&30.32&0.43M \\ \hline
HGN&79&{\bf39.26}&{\bf29.60}&0.43M \\ \hline
\end{tabular}
\label{table:compare with GSH_control_channel}
\small
\centering
\renewcommand\tabcolsep{1pt}
\setlength{\belowcaptionskip}{-2pt}
\caption{Comparison of GraphSH~\cite{2021Graph} and our method using SemGConv and vanilla GConv.}\label{table:compare with GSH}
\subtable[SemGConv]{
\begin{tabular}{l|c|c|c}
\hline
Method &Channels& MPJPE (mm)&\# Params \\ \hline\hline
SeqRes~\cite{zhao2019semantic}&128& 40.78 &0.43M \\ 
GraphSH~\cite{2021Graph} &64& 39.20 & 0.44M \\ \hline
Ours & 64 &38.74 & 0.29M \\
Ours & 128&{\bf 37.32} & 1.04M \\ \hline
\end{tabular}
}
\subtable[Vanilla GConv]{
\begin{tabular}{l|c|c|c}
\hline
Method &Channels&MPJPE (mm)&\# Params \\ 
\hline \hline
SeqRes~\cite{zhao2019semantic}&128& 65.90 &0.30M \\
GraphSH~\cite{2021Graph} &64& 59.10 & 0.22M \\  \hline
Ours & 64 &42.92 & 0.21M \\ 
Ours & 128& {\bf40.66} & 0.71M \\ \hline
\end{tabular}%
}
\end{table}
\begin{figure}[!t]
\renewcommand{\baselinestretch}{1.0}
\setlength{\abovecaptionskip}{-5pt}
\setlength{\belowcaptionskip}{-5pt}
\centering
\vspace{-5pt}
\includegraphics[width=0.4\linewidth]{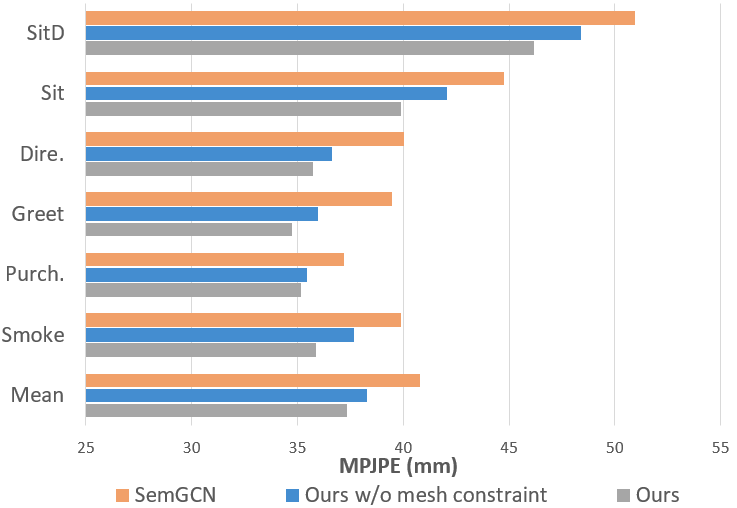}
\includegraphics[width=0.4\linewidth]{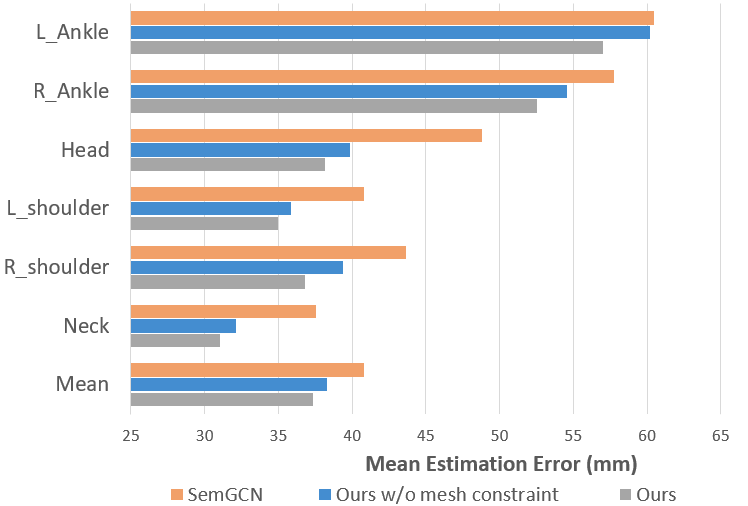}
\caption{Mean estimation errors on different body parts (left) and  actions (right). Ground truth 2D keypoints are used as input. We show the result for \cite{zhao2019semantic} and ours.}\label{fig-action&joint}
\end{figure}

{\bf Understanding the performance improvement.}
We present the average estimation errors of different body parts and actions as well as the overall mean errors in Figure~\ref{fig-action&joint}. Among all the actions, our method obtains larger gains for those with serious self-occlusion, such as \textit{'Sitting'} (4.78mm), \textit{'SittingDown'} (4.85mm), \textit{'Greeting'} (4.74mm), etc, while the overall gain is 3.16mm. For body parts, our method brings much improvement for \textit{'head'} (10.64mm), \textit{'right shoulder'} (6.88mm) and \textit{'left shoulder'} (5.83mm) etc because the vertices of coarse mesh, as shown in Figure~\ref{fig:pipeline}, are denser in the upper part of the human body. Those results prove our opinion that more complex graph structure can bring benefit to the depiction of the human skeleton and some joints with high degrees of freedom.

\begin{table}[!t]
\renewcommand{\baselinestretch}{1.0}
\renewcommand{\arraystretch}{1.0}

\footnotesize
\centering
\setlength{\tabcolsep}{0.75pt}
\renewcommand{\baselinestretch}{1.0}
\renewcommand\arraystretch{1.0}
\setlength{\abovecaptionskip}{-5pt}
\setlength{\belowcaptionskip}{2pt}
\caption{Quantitative evaluation results using MPJPE in millimeter on Human3.6M under Protocol \#1. No rigid alignment or transform is applied in post-processing. \textbf{Detected 2D keypoints} are used as input. $^\dag$ uses temporal information. $^+$ uses extra data from MPII dataset. Best results are highlighted in bold.}\label{table:MPJPE}
\begin{tabular}{lccccccccccccccccc}
\hline
\textbf{Protocol \#1}&Dire.&Disc.&Eat&Greet&Phone&Photo&Pose&Puch.&Sit&SitD.&Smoke&Wait&WalkD&Walk&WalkT&\textbf{Avg.} \\ \hline
Pavllo \etal~\cite{pavllo20193d}$^\dag$&45.2&46.7 &43.3& 45.6& 48.1& 55.1& 44.6& 44.3& 57.3& 65.8& 47.1& 44.0& 49.0& 32.8& 33.9 &46.8\\ 
Cai \etal~\cite{2019Exploiting}$^\dag$ &44.6 &47.4& 45.6& 48.8& 50.8& 59.0&49.7& 47.2& 43.9& 57.9& 61.9&  46.6& 51.3& 37.1& 39.4& 48.8\\ \hline
Martinez \etal~\cite{2017simple} &51.8&56.2&58.1&59.0&69.5&78.4&55.2&58.1&74.0&94.6&62.3&59.1&65.1&49.5&52.4&62.9 \\ 
Pavlakos \etal~\cite{2018Ordinal}$^+$&48.5&54.4&54.5&52.0&59.4&65.3&49.9&52.9&65.8&71.1&56.6&52.9&60.9&44.7&47.8&56.2\\  
Zhao \etal \cite{zhao2019semantic} &48.2&60.8&51.8&64.0&64.6&{\bf53.6}&51.1&67.4&88.7&{\bf57.7}&73.2&65.6&{\bf48.9}&64.8&51.9&60.8\\
Ci \etal~\cite{ci2019optimizing}$^+$&46.8&52.3&{\bf44.7}&{\bf50.4}&{\bf 52.9}&68.9&49.6&46.4&60.2&78.9&{\bf51.2}&50.0&54.8&40.4 &43.3&52.7 \\
Liu \etal~\cite{liu2020comprehensive}  &46.3 &52.2& 47.3& 50.7& 55.5 &67.1& {\bf49.2}&{\bf46.0}& 60.4 &71.1& 51.5& 50.1& 54.5 &40.3& 43.7& 52.4\\
Zou \etal~\cite{zou2020high} &49.0& 54.5& 52.3& 53.6 &59.2& 71.6& 49.6 &49.8& 66.0 &75.5 &55.1 &53.8& 58.5& 40.9 &45.4 &55.6 \\
Xu \& Takano~\cite{2021Graph} &{\bf 45.2}&  {\bf49.9}& 47.5& 50.9& 54.9&  66.1&   48.5&  46.3&  {\bf59.7}&  71.5&51.4&{\bf48.6}&53.9& {\bf 39.9}& 44.1& 51.9 \\ \hline
Ours&47.8&52.5&47.7&50.5&53.9&60.7&49.5&49.4&60.0&66.3&51.8&48.8&55.2&40.5&{\bf42.6}&{\bf 51.8} \\ 
\hline
\end{tabular}
\end{table}
\begin{table}[!t]
\renewcommand{\baselinestretch}{1.0}
\renewcommand{\arraystretch}{1.0}
\footnotesize
\centering
\setlength{\tabcolsep}{0.5pt}
\setlength{\abovecaptionskip}{-8pt}
\setlength{\belowcaptionskip}{2pt}
\caption{Quantitative evaluation results using PA-MPJPE in millimeter on Human3.6M under Protocol \#2. Rigid alignment is applied in post-processing. \textbf{Detected 2D keypoints} are used as input. $^\dag$ uses temporal information. $^+$ uses extra data from MPII dataset. Best results are highlighted in bold.}\label{table:PA-MPJPE}
\begin{tabular}{lccccccccccccccccc}
\hline
\textbf{Protocol \#2    }&Dire.&Disc.&Eat&Greet&Phone&Photo&Pose&Purch.&Sit&SitD.&Smoke&Wait&WalkD&Walk&WalkT&\textbf{Avg.} \\ \hline
Pavllo \etal~\cite{pavllo20193d}$^\dag$&34.2&36.8&33.9&37.5&37.1&43.2&34.4&33.5&45.3&52.7&37.7&34.1&38.0&25.8&27.7&36.8\\ 
Cai \etal~\cite{2019Exploiting}$^\dag$ &35.7&37.8&36.9&40.7&39.6&45.2&37.4&34.5&46.9&50.1&40.5&36.1&41.0&29.6&33.2&39.0\\ \hline 
Martinez \etal~\cite{2017simple} &39.5&43.2&46.4&47.0&51.0&56.0&41.4&40.6&56.5&69.4&49.2&45.0&49.5&38.0&43.1&47.7 \\ 
Pavlakos \etal~\cite{2018Ordinal}$^+$&{\bf34.7}&39.8&41.8&{\bf38.6}&42.5&47.5&38.0&36.6&50.7&56.8&42.6&39.6&43.9&32.1&36.5&41.8\\  
Ci \etal~\cite{ci2019optimizing}$^+$&36.9&41.6&38.0&41.0&41.9&51.1&38.2&37.6&49.1&62.1&43.1&39.9&43.5&32.2&37.0&42.2  \\
Liu \etal~\cite{liu2020comprehensive}  &35.9&40.0&38.0&41.5&42.5&51.4&37.8&36.0&48.6&56.6&41.8&38.3&{\bf42.7}&31.7&36.2&41.2\\
Zou \etal~\cite{zou2020high} &38.6&42.8&41.8&43.4&44.6&52.9&37.5&38.6&53.3&60.0&44.4&40.9&46.9&32.2&37.9&43.7 \\ \hline
Ours&35.8&{\bf 39.7}&{\bf 36.3}&40.6&{\bf 40.2}&{\bf 45.9}&{\bf 36.8} &{\bf 35.8}&{\bf 47.3}&{\bf 53.7}&{\bf 40.7}&{\bf36.4}&43.1&{\bf29.8}&{\bf32.8}&{\bf 39.6}\\ 
\hline
\end{tabular}
\end{table}
\begin{table}[!t]
\small
\centering
\setlength{\tabcolsep}{0.6pt}
\setlength{\abovecaptionskip}{-8pt}
\setlength{\belowcaptionskip}{2pt}
 \caption{Results on the test set of MPI-INF-3DHP~\cite{mehta2017monocular}  by scene.  The results are shown in PCK and AUC.}\label{table:MPI} 
       \begin{tabular}{l|c |ccc|cc}
         \hline
        &Trainning data &~GS~& noGS &Outdoor &ALL (PCK~$\uparrow$) &ALL (AUC~$\uparrow$)\\ \hline \hline
        Martinez \etal~\cite{2017simple} &H36M &49.8&42.5&31.2&42.5&17.0\\
        Mehta \etal~\cite{mehta2017monocular} &H36M&70.8&62.3&58.8&64.7&31.7\\
        Yang \etal~\cite{yang20183d} &H36M+MPII&-&-&-&69.0&32.0\\
        Zhou \etal~\cite{zhou2017towards} &H36M+MPII&71.1&64.7&72.7&69.2&32.5\\
        Luo \etal~\cite{luo2018orinet}  &H36M&71.3&59.4&65.7&65.6&33.2\\
        Ci \etal~\cite{ci2019optimizing} &H36M&74.8&70.8&77.3&74.0&36.7\\
        Zhou \etal~\cite{zhou2019hemlets} &H36M+MPII &75.6&71.3&80.3&75.3&38.0\\
        Xu and Takano~\cite{2021Graph}    &H36M &81.5&81.7&75.2&80.1&45.8\\ \hline
        Ours&H36M& \textbf{87.0}&\textbf{84.9}&\textbf{82.7}&\textbf{85.2}&\textbf{52.1}\\ \hline
        \end{tabular}
\vspace{-10pt}
\end{table}

\subsection{Comparison With the State-of-the-Art}
We use the cascaded pyramid network (CPN)~\cite{chen2018cascaded} as 2D pose detector to obtain 2D input joints for benchmark evaluation. CPN is pre-trained on COCO-dataset and fine-tuned on Human3.6M. Following previous works  \cite{pavllo20193d,zhou2017towards,2019Exploiting}, we also perform horizontal flip augmentation. 
The results are shown in Tables~\ref{table:MPJPE} and~\ref{table:PA-MPJPE} for the two protocols, respectively. Note that some other methods \cite{pavllo20193d,2019Exploiting} that focus on video-based 3D pose estimation are complementary to our method and can be used to improve the performance.

\begin{figure}[!t]
\renewcommand{\baselinestretch}{1.0}
\setlength{\abovecaptionskip}{-10pt}
\centering
\includegraphics[width=0.97\linewidth]{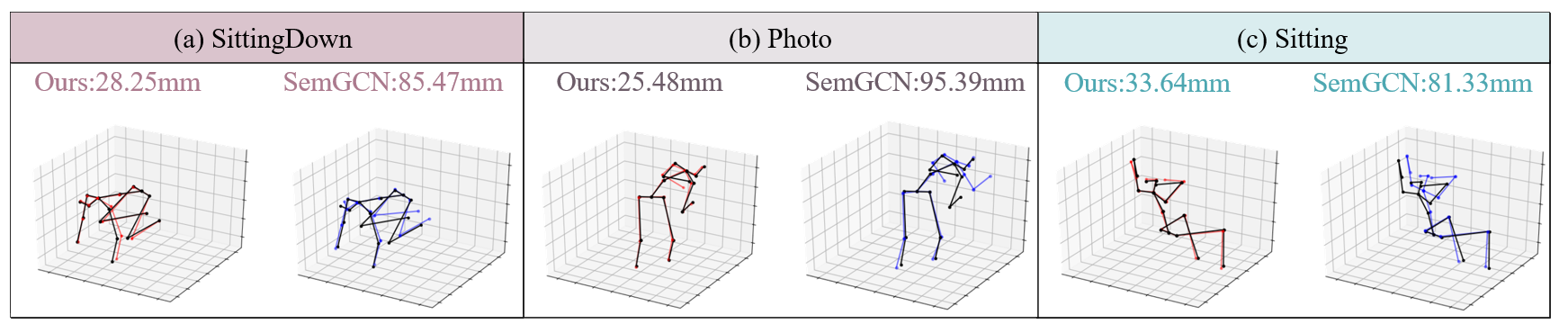}
\caption{Qualitative results of actions with self-occlusion, i.e., SittingDown, Photo, and Sitting. 3D groud truth, ours, and SemGCN are shown in black, red, and blue, respectively.} \label{fig:qualita_compare}
\renewcommand{\baselinestretch}{1.0}
\setlength{\belowcaptionskip}{-10pt}
\centering
\includegraphics[width=1.0\linewidth]{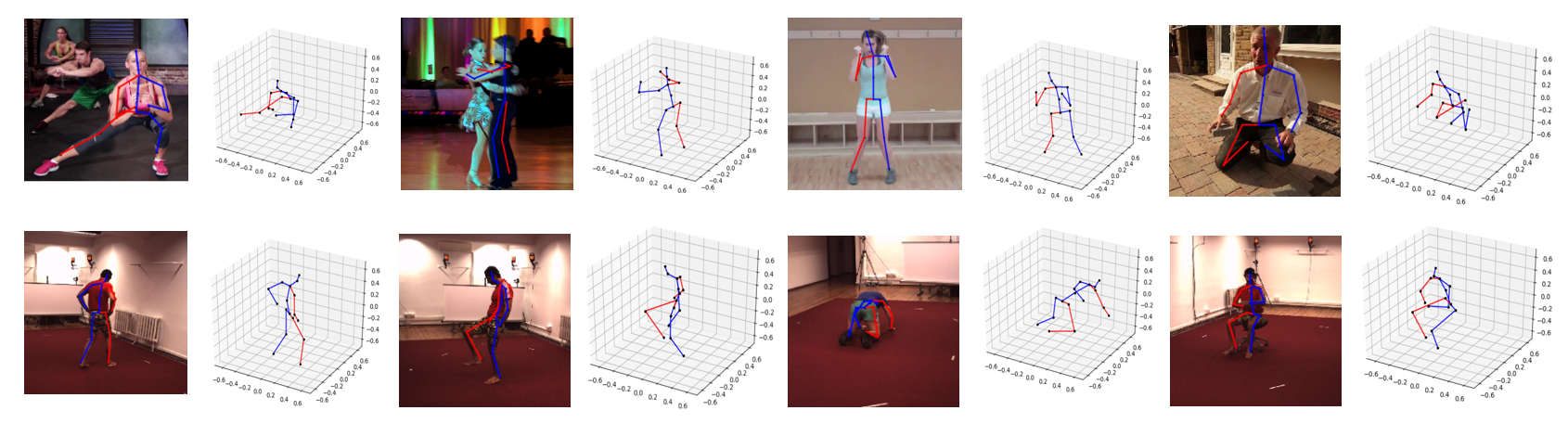}
\caption{Qualitative results of our method on Human3.6M~\cite{ionescu2013human3} (bottom) and MPII~\cite{andriluka20142d} (top).}\label{fig:qualita}
\end{figure}

Tables~\ref{table:MPJPE} and~\ref{table:PA-MPJPE} show that the performance improvement of our model is significant, outperforming  all other GCN-based methods and some representative methods \cite{ci2019optimizing,2018Ordinal} using extra MPII data~\cite{andriluka20142d}. GraphSH~\cite{2021Graph} achieves comparable performance, but it uses more parameters (3.70M) while our model has only 1.04M parameters, and we have proved in Section~\ref{sec:ablation} that our method surpasses GraphSH a lot with the same GCN type and channels. The results demonstrate the great advantage of our HGN. 

Figure~\ref{fig:qualita_compare} shows qualitative results for those actions with serious self-occlusion. Compared with baseline, our HGN can alleviate the depth ambiguity caused by self-occlusion. Figure~\ref{fig:qualita} demonstrates more qualitative results of our HGN on the Human3.6M and MPII datasets. Here, MPII contains in-the-wild images that are unseen for the model trained on Human3.6M. These results further validate the strong generalization ability of our method.
\subsection{Generalization Ability}  
The MPI-INF-3DHP test set~\cite{mehta2017monocular} provides images in three different scenarios: studio with a green screen (GS), studio without green screen (noGS) and outdoor scene (Outdoor). We apply our model to this dataset to test the generalization capabilities of our proposed method and employ 3D-PCK and AUC as evaluation metrics.  As shown in Tab.~\ref{table:MPI}, our model yields 85.2 in PCK and 52.1 in
AUC while only using the Human3.6M dataset for training, which outperforms all the previous state-of-the-arts. These results validate the strong generalization capability of our architecture.

\section{Conclusion}
In this paper, we propose a novel architecture named \textit{Hierarchical Graph Networks (HGN)}. The main contributions are two folds. First, we build a novel sparse-to-fine architecture with multi-scale feature fusion based on the denser graphs generated by a multi-scale graph structure building strategy for better feature extraction. Second, we leverage the human coarse mesh as an additional constraint, refining the local feature representation. Extensive experiment results reveal the benefit of our design. 

\section*{Acknowledgment}
This work was supported in part by the National Natural Science Foundation of China under Grant 61971285, Grant 61871267, Grant 61972256, Grant 61720106001, Grant 61931023, Grant 61932022, Grant 61831018, and in part by the Program of Shanghai Science and Technology Innovation Project under Grant 20511100100.

\bibliography{bmvc_review}
\end{document}